# Machine Learning-based NLP for Emotion Classification on a Cholera X Dataset

Paul Jideani[1, 2][0000-0001-5836-6660] and Aurona Gerber[1,3][0000-0003-1743-8167]

[1] Department of Computer Science, University of the Western Cape, South Africa
[2] Boston City Campus, Stellenbosch, South Africa
[3] Center for AI Research (CAIR), South Africa
`pcijideani@gmail.com`

**Abstract.** Recent social media posts on the cholera outbreak in Hammanskraal have highlighted the diverse range of emotions people experienced in response to such an event. The extent of people's opinions varies greatly depending on their level of knowledge and information about the disease. The documented research about Cholera lacks investigations into the classification of emotions. This study aims to examine the emotions expressed in social media posts about Cholera. A dataset of 23,000 posts was extracted and pre-processed. The Python Natural Language Toolkit (NLTK) sentiment analyzer library was applied to determine the emotional significance of each text. Additionally, Machine Learning (ML) models were applied for emotion classification, including Long short-term memory (LSTM), Logistic regression, Decision trees, and the Bidirectional Encoder Representations from Transformers (BERT) model. The results of this study demonstrated that LSTM achieved the highest accuracy of 75%. Emotion classification presents a promising tool for gaining a deeper understanding of the impact of Cholera on society. The findings of this study might contribute to the development of effective interventions in public health strategies.

## 1 Introduction

Cholera is a waterborne infectious disease that causes severe diarrhoea and vomiting in humans [1, 2]. It remains a persistent menace in many parts of Africa and Asia. Cholera is transmitted by drinking water or eating food contaminated by bacteria [3]. According to the WHO, the disease causes 1.3 to 4 million cases and 21,000 to 143,000 deaths worldwide yearly, mostly in underdeveloped nations, including India (2007), Iraq (2008), Congo (2008), and Zimbabwe (2008-2009) [1, 2, 4]. One of the major concerns of Cholera is its potential to cause large-scale outbreaks that spread rapidly [5, 6]. It has been a source of concern for public health practitioners as outbreaks can overwhelm healthcare systems and strain available resources [7]. The recent cholera outbreak in Hammanskraal, South Africa, has underlined the importance of studying how the general public perceives and responds to the disease. According to the Department of Health, the outbreak started in May 2023, hospitalized 95 people and had a death toll of 31 as of 8 June 2023 [4, 8, 9]. While most research has focused on discovering treatments and vaccinations for the disease, there is growing interest in studying public views and emotions surrounding Cholera, which is the focus of this study [10–12]. This study compared the results of ML-based NLP techniques to analyze public sentiment, specifically the expressed emotions in X data on Cholera.

The remainder of the paper is structured as follows: Section 2 summarises relevant literature. Section 3 covers materials and methods, Section 4 outlines the results and analysis, and Section 5 draws conclusions.

## 2      Literature Review

Public opinion mining, or sentiment analysis, extracts and analyses subjective information from various sources to understand and evaluate public sentiment, attitudes, and opinions towards a particular topic, product, service, or event [13, 14]. It involves the use of natural language processing (NLP) and Machine Learning (ML) techniques to analyze large volumes of textual data, such as social media posts, online reviews, news articles, and customer feedback [15]. Public opinion mining aims to gain insights into how people perceive and feel about a specific subject [16], whilst Sentiment analysis specifically has been used to analyze and classify the sentiment expressed in texts, allowing organizations, businesses, and governments to understand public perception, monitor brand reputation, assess customer satisfaction, predict trends, and make data-driven decisions [17].

Public opinion mining on social media has become increasingly important in today's digital age. Social media platforms have transformed the way people communicate and express their opinions and experiences [18, 19] and organizations use this source of user-generated content to extract insights about public sentiments, attitudes, and opinions, often in real-time [20, 21]. This real-time aspect is vital in today's fast-paced environment because public sentiment can change rapidly. Social media is also a primary platform for sharing information and expressing opinions during a crisis or public event [22]. Public opinion mining, therefore, allows for monitoring public sentiment surrounding the crisis and identifying key issues [23]. These insights are invaluable for making data-driven decisions, refining marketing strategies, staying competitive in the market and even providing input to policymakers and government entities [23–26]. Public opinion mining on social media has implications beyond marketing and business. These insights can guide decision-making processes, help shape policies that align with public expectations, and contribute to better governance [27, 28].

Natural Language Processing (NLP) and Machine Learning (ML) provide powerful techniques for understanding and interpreting human emotions and opinions expressed in text data [29]. NLP enables the processing of textual data, including to tokenize, normalize, and clean the text, remove stop words, handle negation, and identify relevant linguistic features [30, 31]. These pre-processing steps are required to transform the raw text into a format suitable for sentiment analysis. NLP also plays a role in sentiment analysis by utilizing sentiment lexicons or dictionaries. These lexicons contain pre-defined words and their associated sentiment polarities (e.g., positive, negative, or neutral). NLP techniques help match words in the text with entries in the sentiment lexicon, enabling the assignment of sentiment scores to the text [32, 33].

DL models, such as Convolutional Neural Networks (CNNs) and Recurrent Neural Networks (RNNs) have been applied for sentiment analysis [34–36]. These models can learn complex patterns and representations from text data, capturing both local and global dependencies in the text [34]. ML models can be fine-tuned or trained on large-scale datasets, including sentiment-labelled data, to improve their performance in sentiment analysis tasks. Transfer learning techniques enable leveraging pre-trained models on large corpora, improving the efficiency and effectiveness of sentiment analysis models [37, 38]. NLP and DL techniques can also be applied to identify and analyze emotions in sentiment analysis. Emotion detection models can classify text into specific emotional categories, such as happiness, sadness, anger, or fear, providing a more nuanced understanding of sentiment beyond the positive/negative polarity [39].

Several studies documented sentiment analysis concerning COVID-19 and other disease datasets. Jatla and Avula [40] used a Hybrid Deep Sentiment Analysis (HDSA) model to analyze the sentiments in COVID-19-related tweets. The model was trained on a dataset of COVID-19 tweets, achieving a classification accuracy of 94%. Hossain et al. [41]developed a DL-based technique for analyzing COVID-19 tweets, using Bidirectional Gated Recurrent Unit (BiGRU). The model was trained on an improved dataset, and it achieved an accuracy of 87%. Singh et al. [42] conducted a study COVID-19 Twitter data using sentiment analysis and ML techniques to help predict outbreaks and epidemics. The paper discusses various ML techniques such as ME, DT, Support Vector Machine (SVM), and Naive Bayes for sentiment analysis. However, the article does not provide any empirical evidence or case studies to support the effectiveness of the proposed approach. Kaur et al. Field [43] proposed a sentiment analysis DL algorithm for classifying tweets into positive, negative, and neutral sentiment scores using the Hybrid Heterogeneous Support Vector Machine (H-SVM) method.

Ronchieri et al. [44] employed Natural Language Processing (NLP), sentiment analysis, and topic modelling techniques to analyze a dataset comprising 369,472 tweets. The findings revealed prevalent emotions such as fear, trust, and disgust, while prominent discussions centered around topics like malaria, influenza, and tuberculosis. The analytical methods applied included the Latent Dirichlet Allocation (LDA) model and TF-IDF vectorization.

Han & Thakur [45] performed sentiment analysis using a methodology that involved analysing 12,028 tweets focusing on the Omicron variant. The results indicated that 50.5% of the tweets conveyed a neutral sentiment, and the predominant language used was English, accounting for 65.9%. Oladipo et al. [46] presented a study wherein sentiment analysis was conducted using the NRC lexicon approach. The results revealed an overall positive sentiment towards the lockdown exercise, based on an analysis of 22,249 tweets sourced from national stakeholders and the general public. Shaalan et al. [47] presented a sentiment analysis study focused on evaluating sentiments expressed in Arabic tweets related to COVID-19. The study employed various models, starting with data acquisition followed by text pre-processing. Term Frequency Inverse Document Frequency (TF-IDF) was utilized to generate feature vectors. Multiple classifiers, including Naïve Bayes, Support Vector Machine, Logic Regression, Random Forest, and K-Nearest Neighbour, were compared through experiments. Performance evaluation was conducted using metrics such as Precision, Accuracy, Recall, and F1 Score. The most effective model achieved an accuracy of approximately 84%.

As is indicated in the summary of related work above, the use of NLP and ML for opinion mining to detect emotions in Twitter (X) data is a well-recognized technique, and this study adopted this approach to detect emotions regarding the Cholera outbreak in South Africa.

## 3      Materials and Methods

The framework in Figure 1 provides an overview of the research methodology, consisting of dataset management and the application of DL algorithms. Dataset management consists of dataset collection, pre-processing, and balancing. These stages are fundamental in ensuring the quality, consistency, and unbiased representation of the data. The application of ML algorithms presents the specific algorithms employed: Bidirectional Encoder Representations from Transformers (BERT), Long Short-Term Memory (LSTM), Logistic Regression, and Decision Tree.

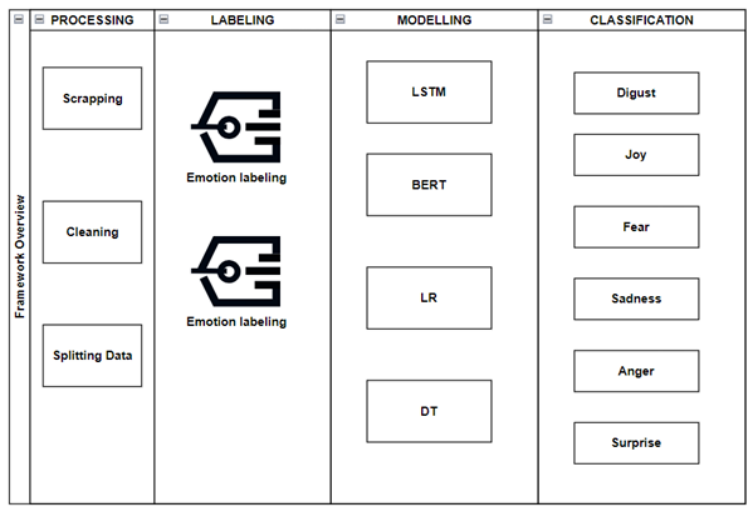

**Fig. 1.** Research Methodology Overview

- **Dataset Collection:** The Twitter API was used to extract tweets using #Hammanskraal # CholeraOutbreak, #water, #Cholera hashtags during a specific time range using the Tweepy library. The information extracted from each returned tweet included the tweet ID, text, and creation date. Between April 2023 and July 2023, a total of 23000 tweets were collected. The script included monitoring statements to print the current date and the total number of collected tweets. After completing the scraping, a Pandas DataFrame was generated from the list of tweets, which was then saved to a CSV file.

- **Dataset Pre-processing:** The collected dataset was pre-processed for sentiment analysis. A tokenizer was trained on the complete dataset. This process involved the tokenizer learning the vocabulary and word-to-index mappings from the provided data. Next, the trained tokenizer was utilized to transform both the training and test data into sequences of numerical indices, representing the words. This conversion enabled the textual data to be represented in a format that the ML algorithms can process. Sequences were padded to ensure uniform lengths, which is necessary for input consistency in neural network models. Finally, the labels for both the training and test sets underwent transformation into dummy variables. This transformation (or one-hot encoding) converted categorical labels into binary vectors, making them suitable for training machine learning models. Furthermore, the tweets in the file were thoroughly cleaned by eliminating duplication, extending contractions, removing stopwords, stripping numeric values, stripping punctuation, removing extra spaces, and deleting greetings and other unimportant words.

- **Emotion Labelling:** The pysentimiento library was used to analyze emotion within tweets. Utilizing Ekman's basic wheel of emotions, the objective was to discern six specific emotions within a corpus of tweets. The sentiment with the highest probability was extracted from the predicted emotions, excluding the "others" category, and appended to the predicted emotions list. Throughout the iteration, the output was cleared and the current number of predicted emotions was printed to monitor the progress. Once all tweets were analyzed, the predicted emotions the results were saved as a CSV file. Algorithm 1 illustrates the emotion labelling algorithm.

---

**Algorithm 1 LABELING STEPS**

---

```
Start:
    # Iterate through each row of the DataFrame
    for each row in emo_analysis_data1:
        # Predict emotion using the emotion analyzer on the 'Text' column
        output = emotion_analyzer.predict(row['Text'])

        # Extract the sentiment with the highest probability, excluding
"others"
        sentiments = [sentiment for sentiment in output.probas.keys() if
sentiment != "others"]
        highest_sentiment = max(sentiments, key=lambda sentiment: out-
put.probas[sentiment])

        # Append the highest sentiment to the predicted_emotions list
        predicted_emotions.append(highest_sentiment)

    # End emotion analysis process
End:
```

---

```
    # Add the predicted_emotions list as a new column called 'Emotion' to
the DataFrame
    emo_analysis_data1['Emotion'] = predicted_emotions
Disgust, Joy, Fear, Sadness, Anger, and Surprise were the types of emo-
tions detected in our dataset. As shown in Figure 4 Digust was the most
prevalent emotion with 38.8% of all emotions available, while Surprise was
the least expressed emotion with 1.8%.
```

**Dataset Balancing:** The data was balanced for sentiment analysis using a two-step approach. Initially, the dataset was split into training and test sets. Within the training set, both majority and minority classes were identified based on the target variable. The minority class underwent an oversampling process to address the class imbalance. This involved generating additional instances of the minority class to balance its representation within the dataset. By utilizing the resample() function from the scikit-learn library, instances from the minority class were duplicated until its size equalled that of the majority class. Subsequently, the upsampled minority class was combined with the original majority class, resulting in a more balanced training dataset. Finally, the effectiveness of the oversampling technique was confirmed by examining the new distribution of class counts.

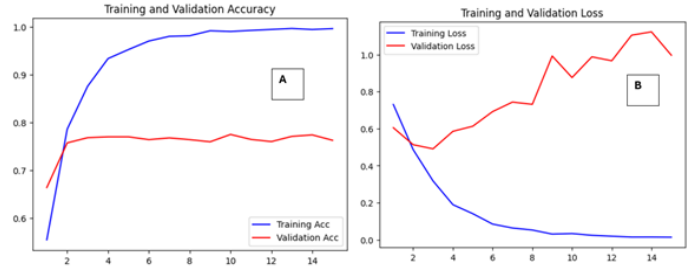

**Fig. 2.** Training and Validation Accuracy

Figures 2 A & B provide a line plot using the Matplotlib library to visualize the training and validation accuracy across various epochs during the training of a machine learning model. The x-axis denotes the epochs, which represent the iterations of the training process. On the y-axis, the accuracy values are depicted. The training accuracy is depicted in blue, while the validation accuracy is shown in red. The legend helps distinguish between the two lines, specifying which corresponds to the training accuracy and which to the validation accuracy. The plot is then displayed using the plt.show() function.

### 3.1 Machine Learning Algorithms

This study analyzed four DL models: LSTM, Logistic Regression, Decision Tree, and BERT. A brief description of the four algorithms follows.

1. **Long Short-Term Memory (LSTM):** LSTMs address the vanishing gradient problem in traditional RNNs by introducing gating mechanisms. These mechanisms allow LSTMs to selectively retain and forget information over long periods [48]. As a result, LSTMs are better at capturing long-term dependencies, which is essential for many NLP tasks, such as language modelling and machine translation [49]. The LSTM (Long-Term Memory) model architecture in this study is composed of three LSTM layers and two fully connected layers. Each LSTM layer is configured with an embedding layer having a dimension of 64, an input length of 29, and a vocabulary size of 100,000. The choice of the specific input length and vocabulary size depends on the characteristics of the dataset, and these values need to be filled in based on the actual data. The activation function employed for each LSTM layer is the Rectified Linear Unit (ReLU), known for introducing non-linearity to the model. Additionally, each LSTM layer is followed by a dropout layer, a regularisation technique designed to prevent overfitting by randomly dropping a certain percentage of connections during training.

2. **Bi-directional Encoder Representations from Transformers (BERT):** BERT is a pre-trained transformer-based language model that captures bi-directional context from text. It is trained on massive text data and learns to generate deep contextualized word embeddings [50]. These contextualized embeddings have been proven to be highly effective in various downstream NLP tasks. BERT has achieved remarkable results in tasks like question-answering, natural language inference, and named entity recognition [51]. The key idea behind BERT is bi-directionality, which allows the model to consider the entire context of a word by looking both to the left and right of it. Unlike traditional language models that process text in a unidirectional manner, BERT is pre-trained using a masked language modelling objective [52]. During pre-training, some of the words in a sentence are masked, and the model is trained to predict the masked words based on the surrounding context. This process enables BERT to generate deep contextualized word embeddings that capture complex semantic relationships between words in a sentence [53].

   The BERT model architecture in this study comprises multiple transformer layers designed for bidirectional context representation. Unlike traditional sequential models, BERT operates bidirectionally, simultaneously capturing contextual information from both left and right directions. In this specific implementation, the BERT model consists of several transformer layers, with the exact number not specified. Each transformer layer is equipped with self-attention mechanisms, allowing the model to weigh different parts of the input sequence differently based on their relevance to each other. This attention mechanism enables BERT to effectively capture long-range dependencies and contextual information [54]. Like the LSTM model, BERT uses an embedding layer to represent input tokens. However, in BERT, the embeddings are contextualized, meaning they are generated considering the entire input sequence rather than in isolation.

3. **Logistic Regression (LR):** The Logistic Regression model is a simple yet powerful linear model that is widely used in binary classification tasks [55]. It is a probabilistic model that predicts the probability of an instance belonging to a specific class [56]. Logistic Regression can efficiently handle feature interactions and provide interpretable results, making it a popular choice for sentiment analysis.

4. **Decision Tree:** Decision Trees are a non-parametric supervised learning method that learns a hierarchical structure of if-else rules to make predictions [57]. Decision Trees can handle both categorical and numerical features and capture complex feature interactions. They can also provide interpretable rules, making them suitable for sentiment analysis tasks where it is important to understand the reasoning behind the predictions [58, 59].

The rationale behind using the four algorithms for emotion prediction can be attributed to their unique characteristics and advantages in handling different aspects of sentiment analysis tasks. LR can efficiently handle feature interactions and provide interpretable results, while Decision Trees can handle both categorical and numerical features and capture complex feature interactions. BERT is a state-of-the-art transformer-based model that can capture intricate linguistic patterns and nuances, making it highly effective for sentiment analysis [60, 61]. LSTM and BERT are particularly adept at capturing contextual information, while Logistic Regression and Decision Tree provide simpler and more interpretable models. This allows for evaluation results on which model works best for a specific sentiment analysis task based on accuracy, interpretability, computational efficiency, and resource requirements.

### 3.2    Experimental Setup and Performance Metrics

Different experiments were performed to evaluate the effectiveness of the four models that were developed in this study. The models were implemented with the Keras Library and Python. The cleaned dataset consisted of 19077 tweets, which is divided into 80% for training (15262 instances) and 20% for testing (3815 instances). The models were trained on 80% of the dataset and evaluated on the remaining 20%.

The assessment of the model performance was achieved through the utilization of performance evaluation metrics. Numerous metrics have been introduced in documented research, each focusing on specific facets of algorithmic performance. Hence, for every machine learning problem, a suitable set of metrics is essential for accurate performance evaluation. In this study, we employ several standard metrics commonly used for classification problems to derive valuable insights into algorithm performance and facilitate a comparative analysis. This study adopted four performance metrics: accuracy, precision, recall, and F1 score. The metrics can be calculated using equations (1) - (4).

$$Precision = \frac{TP}{TP + FP} \quad (1) \qquad\qquad Recall = \frac{TP}{TP + FN} \quad (2)$$

$$F1 - score = 2 * \frac{Precsion * Recall}{Precision + Recall} \quad (3) \qquad Accuracy = \frac{TP + TN}{TP + TN + FP + FN} \quad (4)$$

## 4      Results and Analysis

Experiments were conducted to assess the effectiveness of LSTM, DT, LR and BERT. Additionally, a comparative analysis is provided to highlight the relative performance of these four models.

Figure 3 and Figure 4 depict the resulting distribution of emotions based on Ekman's Basic Wheel in the X cholera datasets. Notably, 691 instances conveyed 'anger,' reflecting discontent, possibly because the authorities handled the Cholera crisis. A substantial number of tweets, totalling 5822, expressed 'disgust' towards the unfolding cholera outbreak. 'Fear' was evident in 3172 tweets, signifying anxiety about the impact of the Cholera outbreak. On the positive spectrum, 4192 tweets suggested potential optimistic developments or an improving situation. The emotion of 'sadness' emerged in 853 tweets reflecting the adverse effects of the cholera outbreak on individuals. Lastly, 270 tweets conveyed 'surprise,' at the occurrence of a cholera outbreak in the modern age.

### 4.1      Performance Analysis

As shown in Table 1, LSTM achieved a classification accuracy of 76%. The algorithm correctly predicted the correct class labels for a significant portion of the dataset. Also, LSTM produced a precision, recall, and F1-score of 75%, 81%, and 78%, respectively, implying that  LSTM achieved a high overall accuracy of 76% and exhibited a notable balance between precision, recall, and F1-score. The precision of 75% indicates that among the instances predicted as positive, 75% were indeed true positives. The recall of 81% suggests that the model successfully captured a significant proportion of actual positive instances. The F1-score of 78%, which is the harmonic mean of precision and recall, further consolidates LSTM's performance by considering both false positives and false negatives. These metrics collectively affirm the robustness of LSTM in effectively classifying instances, striking a commendable balance between avoiding false positives and capturing true positives in the classification process.

Results indicate that LR produced a classification accuracy of 60%. LR correctly classified 60% of the emotions in the dataset. The precision of 59% signifies that, among the instances predicted as positive, 59% were indeed true positives. On the other hand, the recall of 86% suggests that LR successfully captured a substantial proportion of actual positive instances. The F1-score of 70%, the harmonic mean of precision and recall, provides a balanced assessment by considering false positives and false negatives. While LR demonstrated high recall, indicating its ability to identify positive instances, the precision was relatively lower, implying a higher rate of false positives. This underlines LR's potential for improving the trade-off between precision and recall.

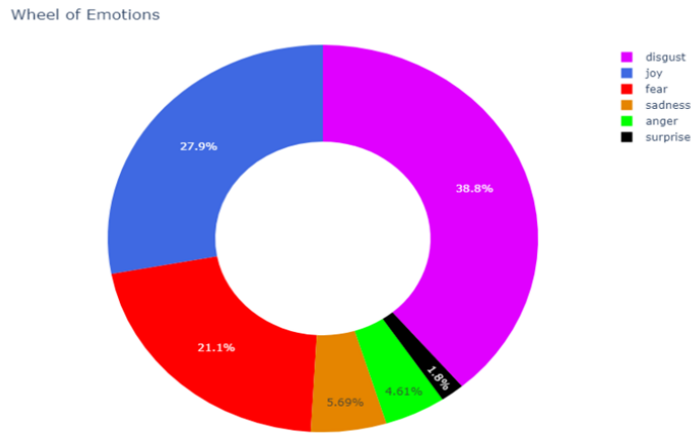

**Fig. 3.** Wheel of emotions expressed from the dataset

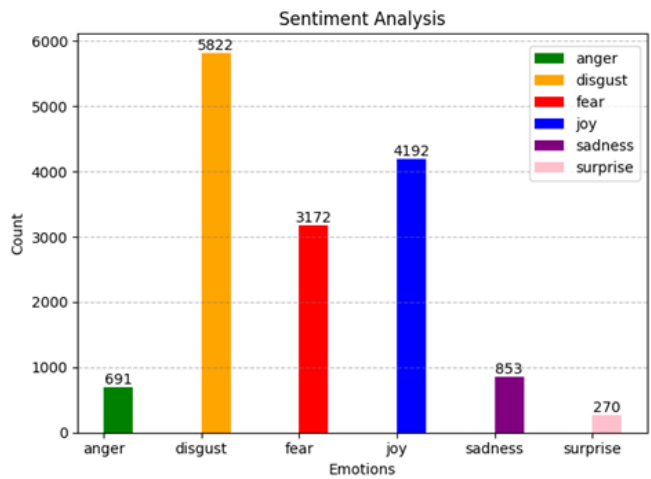

**Fig. 4.** Visualization of the number of emotions

The results show that the Decision Tree (DT) produced a classification accuracy of 56%. While accuracy provides an overall view of the model's performance, further insights can be gained by examining precision, recall, and F1-score. The precision of 59% indicates that among instances predicted as positive by DT, 59% were true positives. Simultaneously, the recall of 86% suggests that DT effectively captured a substantial proportion of actual positive instances. The F1-score, which combines precision and recall, stands at 70%, offering a balanced evaluation. Notably, the performance metrics for DT closely resemble those of Logistic Regression, indicating a comparable ability to identify positive instances. However, similar to LR, the precision-recall trade-off should be considered for potential refinements in DT's performance.

BERT achieved a classification accuracy of 66%. While accuracy is a fundamental metric, a more comprehensive evaluation involves examining precision, recall, and F1-score. The precision of 65% indicates that among instances predicted as positive by BERT, 65% were true positives. Simultaneously, the recall of 68% suggests that BERT effectively captured a substantial proportion of actual positive instances. The F1-score, combining precision and recall, stands at 67%, offering a balanced measure of BERT's performance. The results demonstrate that BERT exhibits effectiveness in sentiment analysis on the given dataset. Further, improved analyses, such as exploring potential biases or examining performance across different emotion classes, could provide additional insights into BERT's performance.

**Table 1.** Results for class 0 – majority class

| Model | Accuracy | Precision (%) | Recall (%) | F1-score (%) |
|-------|----------|---------------|------------|--------------|
| LSTM  | 76       | 75            | 81         | 78           |
| LR    | 60       | 59            | 86         | 70           |
| DT    | 56       | 59            | 86         | 70           |
| BERT  | 66       | 65            | 68         | 67           |

**Table 2.** Results for class 1 – minority class

| Model | Accuracy | Precision (%) | Recall (%) | F1-score (%) |
|-------|----------|---------------|------------|--------------|
| LSTM  | 76       | 74            | 66         | 70           |
| LR    | 60       | 60            | 26         | 36           |
| DT    | 56       | 60            | 26         | 36           |
| BERT  | 66       | 66            | 63         | 65           |

Despite limitations, the study's findings offer insights into public sentiments and emotions related to cholera outbreaks as expressed on social media platforms.

### 4.2 Comparative Performance Analysis

Among the four models evaluated, LSTM had the highest accuracy and better precision, recall, and F1-score performance for both classes. Logistic Regression and Decision Tree models show similar results with the lowest accuracy and weaker performance in predicting the minority class (class 1). BERT offers a middle ground with moderate accuracy and relatively balanced precision, recall, and F1-scores. LSTM has the highest overall accuracy of 75%, and performs relatively well in terms of precision, recall, and F1-score for both classes.

The models have varying precision and recall values for different classes, which provide insights into their performance on each class. LSTM has balanced precision and recall values for both classes, indicating that it predicts both negative (class 0) and positive (class 1) instances well. Logistic Regression and Decision Tree have relatively high recall for class 0 but lower recall for class 1. This suggests they are better at identifying negative instances (class 0) than positive ones (class 1). The precision for class 1 is notably lower, implying that when they predict class 1, they tend to have a higher rate of false positives. BERT shows balanced precision and recall values for both classes, similar to the LSTM model. This indicates that it performs reasonably well in identifying both negative and positive instances.

LSTM has relatively high F1-scores for both classes, indicating its ability to balance precision and recall for both negative and positive instances. Logistic Regression and Decision Tree these models have lower F1-scores for class 1, suggesting challenges in effectively predicting positive instances (class 1) due to lower precision and recall. The F1-scores of BERT are balanced for both classes, indicating a balanced performance in terms of precision and recall for both negative and positive instances.

The evaluation metrics for the Logistic Regression and Decision Tree models are identical, suggesting comparable performance between the two. Nonetheless, these models exhibit lower accuracy (0.60) in contrast to the LSTM model. On the other hand, the BERT model showcases a respectable accuracy score of 0.66. Notably, BERT also demonstrates balanced precision, recall, and F1-scores for both classes. Considering these factors, the chosen approach is to proceed with the LSTM model due to its higher accuracy and performance across various evaluation criteria. The sentiment analysis conducted on the cholera X dataset through machine learning models has yielded findings regarding the emotional responses of individuals to the Hammanskraal cholera outbreak. The Long short-term memory (LSTM) model emerges as a standout performer, demonstrating a commendable ability to balance precision and recall for both negative and positive sentiments. This suggests that LSTM effectively captures the nuances of sentiment expressed in tweets related to the cholera situation, establishing it as a robust model for sentiment analysis in this context. On the other hand, Logistic Regression and Decision Tree models face challenges, particularly in effectively predicting positive sentiments. The lower F1-scores for class 1 (positive sentiment) indicate that these models may struggle to capture the optimistic or supportive sentiments related to the cholera outbreak. Further exploration and feature refinement might be necessary to enhance the performance of these models. In contrast, the Bidirectional Encoder Representations from Transformers (BERT) model exhibits a balanced performance with equally high F1-scores for both negative and positive sentiments. This balanced performance underscores BERT's comprehensive understanding of the diverse emotions expressed in tweets related to the cholera outbreak. BERT's ability to capture the complexity and variability of sentiment positions it as a suitable model for studying sentiments in this domain.

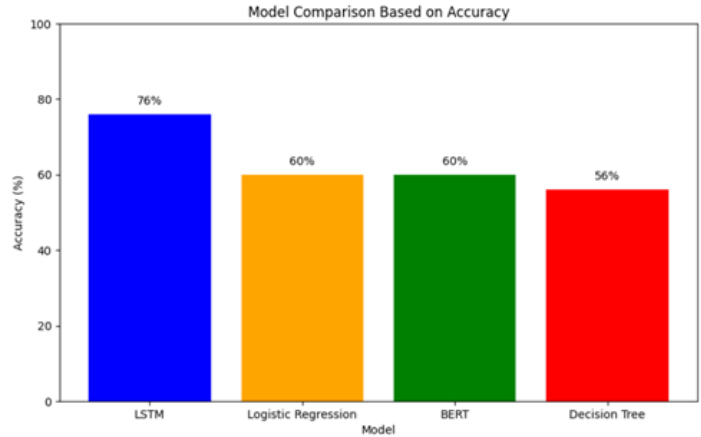

**Fig. 5.** Model performance based on accuracy

## 5    Conclusion

The emotion classification of social media text data connected to disease outbreaks contributes to a better understanding of people in such situations. Our study classified cholera tweets based on their emotional content. We extracted over 23000 tweets from X across several languages. The dataset was pre-processed before being labelled with NLTK's sentiment analyzer and separated into 80% training and 20% test data. Six (6) emotions as suggested by  [62] were identified in our classification. Moreover, ML and DL models were developed for emotion categorization, using LSTM, LR, DT, and BERT. The models were trained on the cholera dataset were LSTM, LR, DT, and BERT produced a classification accuracy of 76%, 60%, 56%, and 66% respectively.

The emotion classification results show that the sentiment of tweets is predominantly negative (6,307 tweets).  4,790 tweets expressed a positive sentiment and 3,903 tweets were classified as neutral. Using Ekman's basic emotions, the finding of the study reveals disgust as the dominant emotion expressed, accounting for a significant emotion observed in the dataset. Various factors can influence this sentiment distribution, such as: (i) Public Perception: Cholera, which is one of the hashtags used to scrape tweets, is a serious and potentially life-threatening disease. Negative sentiments may arise from fear, concern, or negative experiences related to Cholera and its impact on individuals, communities, or public health. (ii) Outbreak Context: The occurrence of a cholera outbreak can amplify negative sentiments. During an outbreak, there may be heightened public attention, media coverage, and discussions focused on the negative aspects, such as the spread of the disease, its impact on affected areas, and the challenges in controlling and managing the outbreak. (iii) Emotional Impact: Diseases like Cholera can evoke strong emotional responses, particularly when they affect vulnerable populations or regions with limited healthcare infrastructure. Negative sentiments may arise from empathy towards affected individuals, frustration with handling the outbreak, or anger towards perceived negligence or inadequate response measures. Considering these factors and interpreting the sentiment results in the context of the specific dataset and analysis methodology used is important.

This provides a mechanism to gain insight into public perception and could be used for real-time monitoring of public sentiment during events such as Cholera outbreaks. Misinformation and rumours related to the outbreaks could also be detected, which is crucial for maintaining public trust and providing accurate information. Future research could explore integrating different data modalities, such as images, videos, and user engagement metrics, with text-based sentiment analysis. Analyzing multiple modalities can offer a more comprehensive understanding of emotions and sentiments related to health crises like cholera outbreaks. Furthermore, conducting longitudinal studies over multiple cholera outbreaks can reveal temporal patterns of emotions and sentiments. Investigating how emotions evolve over time can provide insights into public reactions at different stages of the outbreak and aid in developing targeted interventions.

**Disclosure of Interests.** The authors have no competing interests to declare that are relevant to the content of this article.